\documentclass[runningheads]{llncs}
\usepackage{eccv}
\usepackage{colortbl}
\usepackage{amsmath}
\usepackage{gensymb} 
\usepackage{booktabs} 
\usepackage{graphicx} 
\usepackage{multirow} 

\definecolor{tabfirst}{rgb}{1,0.8,0.8}
\definecolor{lightgreen}{rgb}{0.8,1,0.8}
\definecolor{headerblue}{rgb}{0.8,0.8,1}
 
\usepackage{eccvabbrv}

\usepackage{hyperref}

\usepackage{orcidlink}

\begin{document}

\title{Invertible Neural Warp for NeRF}


\author{Shin-Fang Chng\and
Ravi Garg  \and
Hemanth Saratchandran  \and
Simon Lucey}


\authorrunning{Shin-Fang et al.}

\institute{Adelaide University \\
Australian Institute for Machine Learning \\
\email{shinfang.chng@adelaide.edu.au}\\
\url{https://sfchng.github.io/ineurowarping-github.io/}}

\maketitle

\begin{abstract}


This paper tackles the simultaneous optimization of pose and Neural Radiance Fields (NeRF). Departing from the conventional practice of using explicit global representations for camera pose, we propose a novel overparameterized representation that models camera poses as learnable rigid warp functions. We establish that modeling the rigid warps must be tightly coupled with constraints and regularization imposed. Specifically, we highlight the critical importance of enforcing invertibility when learning rigid warp functions via neural network and propose the use of an Invertible Neural Network (INN) coupled with a geometry-informed constraint for this purpose. We present results on synthetic and real-world datasets, and demonstrate that our approach outperforms existing baselines in terms of pose estimation and high-fidelity reconstruction due to enhanced optimization convergence.




\end{abstract}

\section{Introduction}
\label{sec:intro}

\begin{figure}[t]
    \centering
    \begin{subfigure}[t]{0.18\textwidth}
        \includegraphics[width=\linewidth]{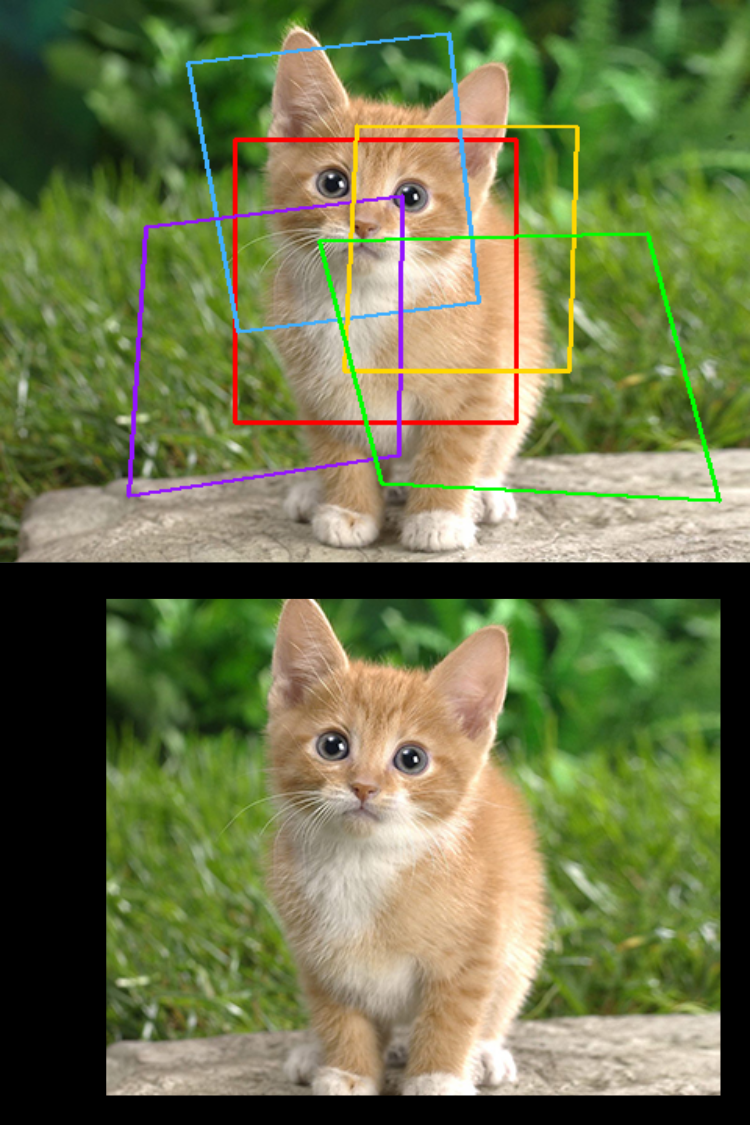}
        \captionsetup{justification=centering}
        \caption{Groundtruth}
        \label{fig:planar_exp_gt}
    \end{subfigure}
    \begin{subfigure}[t]{0.18\textwidth}
        \includegraphics[width=\linewidth]{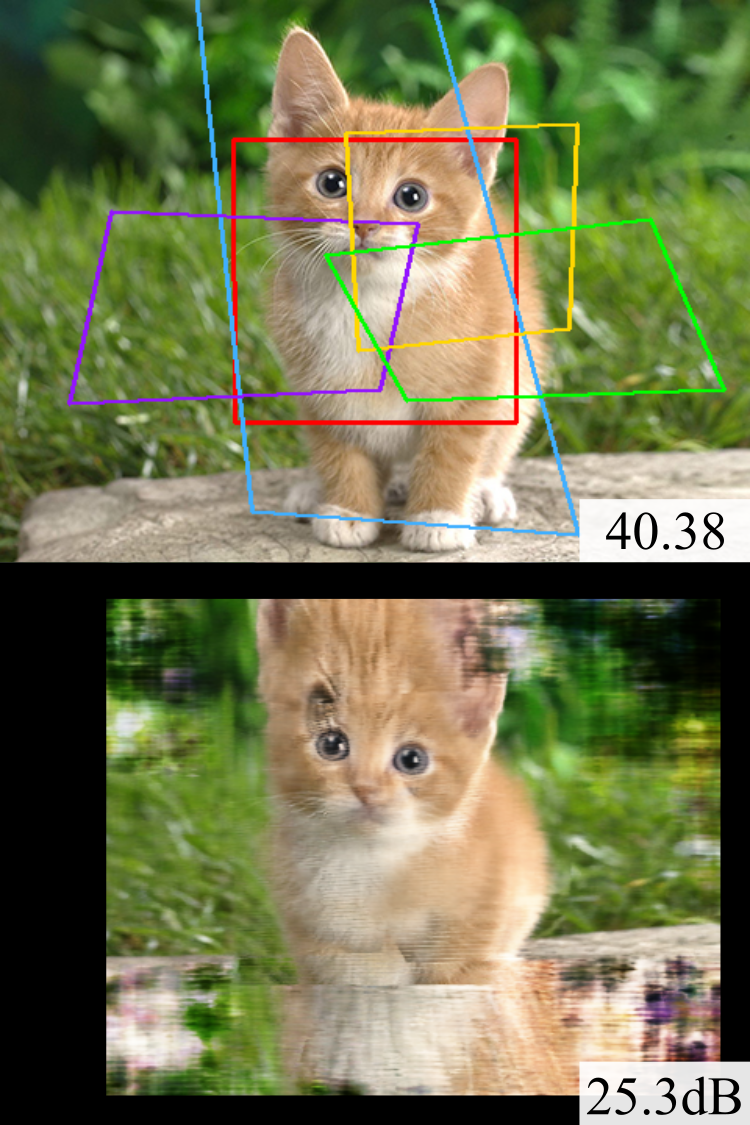}
        \caption{BARF~\cite{lin2021barf}}
        \label{fig:planar_exp_barf}
    \end{subfigure}
    \begin{subfigure}[t]{0.18\textwidth}
        \includegraphics[width=\linewidth]{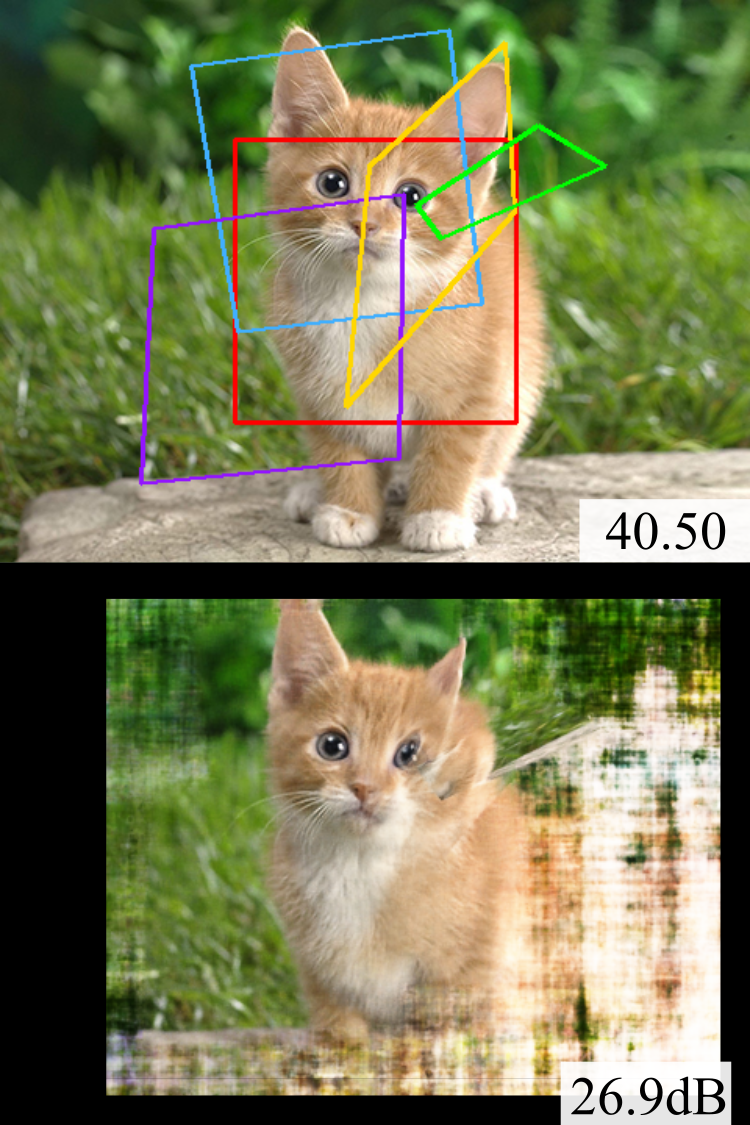}
        \caption{Naive}
        \label{fig:planar_exp_mlp}
    \end{subfigure}
    \begin{subfigure}[t]{0.18\textwidth}
        \includegraphics[width=\linewidth]{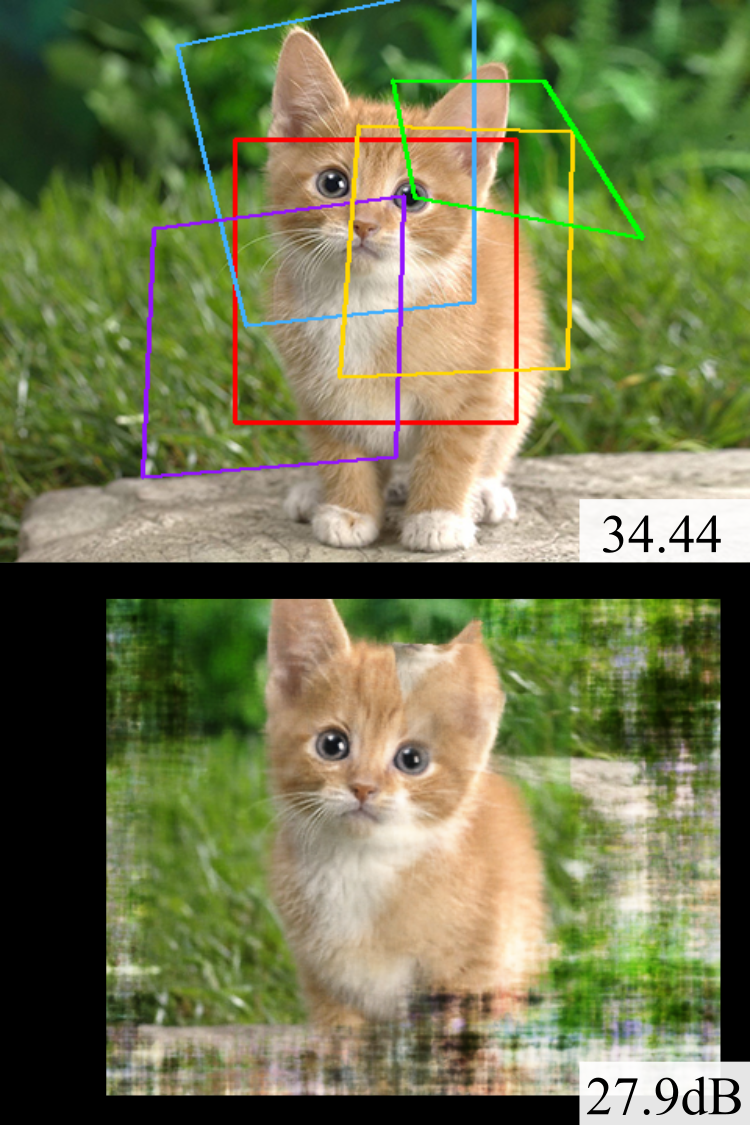}
        \captionsetup{justification=centering}
        \caption{Implicit-Invertible MLP}
        \label{fig:planar_exp_cmlp}
    \end{subfigure}
    \begin{subfigure}[t]{0.18\textwidth}
        \includegraphics[width=\linewidth]{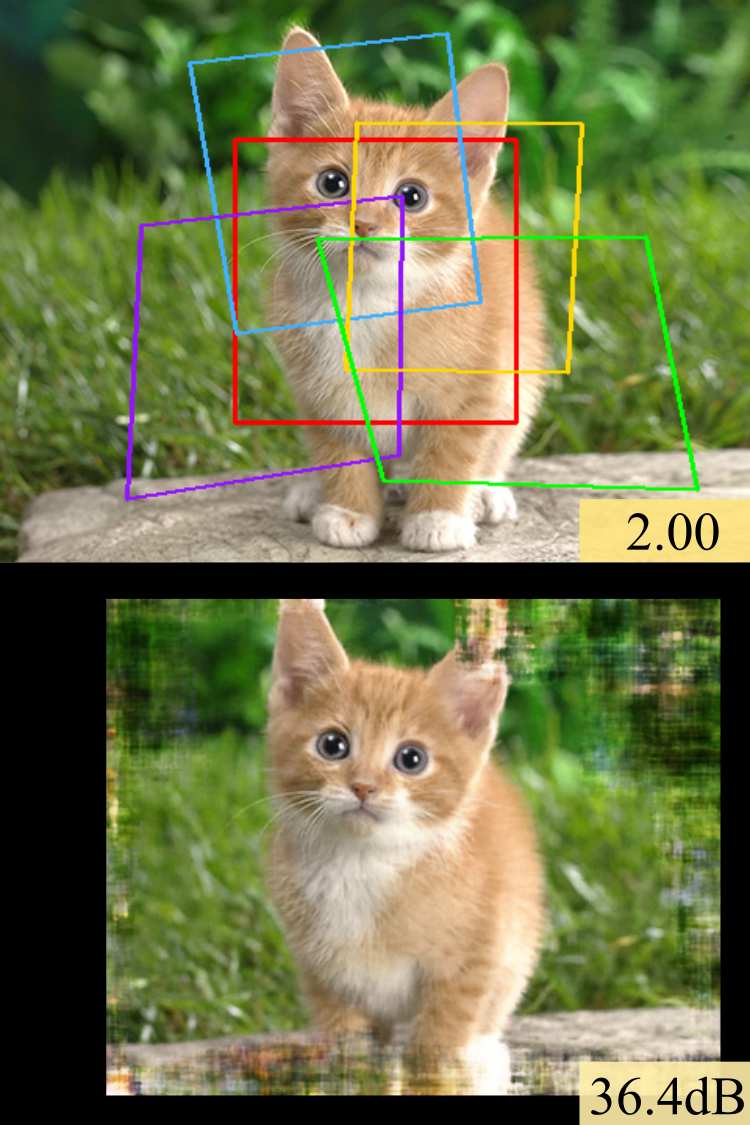}
        \captionsetup{justification=centering}
        \caption{Explicit-Invertible INN}
        \label{fig:planar_exp_inn}
    \end{subfigure}
    \vspace{-1em}
    \caption{We investigate how overparameterizing rigid warps of rays with an MLP benefits the joint optimization task of camera pose and NeRF. This example estimates the warps that align the color-coded patches in~\cref{fig:planar_exp_gt} while solving for the neural field. Unlike ``BARF'' and ``naive'' MLP pose overparameterization methods fail catastrophically, enforcing invertibility, either \textit{implicitly}~(\cref{fig:planar_exp_cmlp}) or \textit{explicitly}~(\cref{fig:planar_exp_inn}) significantly improves warp estimation, see \cref{subsec:exp_baselines} for details of each method. We establish that \textit{invertibility} is a crucial for MLP-based rigid warp representation.}
    \vspace{-1em}
    \label{fig:planar_exp}
\end{figure}
NeRF~\cite{mildenhall2021nerf} has recently emerged as a compelling approach for synthesizing photorealistic images from novel views. NeRF employs a multi-layer perceptron (MLP) to model a volumetric representation of a 3D scene. It operates by minimizing the photometric loss, which is the discrepancy between rendered images and actual images. NeRF's ability to reconstruct high-fidelity signals, coupled with its memory efficiency, has propelled its adoption across a wide array of applications \cite{zhu2022nice,park2021nerfies,zhan2022activermap,zhao2022humannerf,guo2020object,xu2023nerf,xu2022point, kerr2023lerf}, demonstrating its significant impact and versatility.

One of the primary challenges with NeRF is the requirement for precisely known camera poses for each captured image. To address this challenge, several approaches such as BARF \cite{lin2021barf}, NeRFmm \cite{wang2021nerf}, and GARF \cite{chng2022gaussian} have been developed. These methods facilitate the simultaneous optimization of the NeRF and the camera poses, using a compact, six-dimensional vector to represent the camera poses efficiently. However, this compact parameterization, while prevalent in contemporary structure from motion (SfM) literature~\cite{schoenberger2016sfm,parra2021rotation}, has been shown to struggle with poor basin of convergence when solved simultaneously with a NeRF~\cite{chen2023local}. Drawing wisdom from machine learning, where overparameterization has been recognized as a catalyst for enhanced optimization convergence in modern deep neural networks \cite{li2018learning, srivastava2014dropout, neyshabur2015norm}, this paper explores the potential of pose overparameterization for simultaneous pose and neural field estimation.


\noindent \textbf{Our approach:} In traditional NeRF setups, accurately known extrinsic camera pose, comprising of a global rotation and translation for each image, are used to explicitly map pixel coordinates and the camera center to determine the viewing rays in a global world coordinate system~\cite{mildenhall2021nerf}. Following the warping operation, the colors and volume densities along each ray in the world coordinate space is manipulated individually through a photometric loss function. In this paper, we explore scenarios where camera poses are not known. Specifically, we propose using a neural network to model the \textit{rigid warp function} of ray. While it may seem counterintuitive to replace a succinct pose function with a more complex MLP, we argue that the enhanced convergence properties of such overparameterization \cite{nguyen2020global,neyshabur2015norm, allen2019learning} -- in conjunction with the right constraint and prior -- outweigh the increased functional flexibility. 

Additionally, we highlight the critical role of enforcing \textit{invertibility} when learning rigid warps using an MLP.~\footnote{Our use of invertibility strictly adheres to the well-established mathematical definition. Let $f$ be a function whose domain is $\mathcal{X}$ and codomain is $\mathcal{Y}$. $f$ is invertible iff there exists a function $g$ from $\mathcal{Y}$ to $\mathcal{X}$ such that $g(f(x))=x$ $\,\forall x \in \mathcal{X}$ and $f(g(y))=y$ $\,\forall y \in \mathcal{Y}$~\cite{jeffreys1999methods}. We use bijective and invertible interchangeably throughout our paper. } 
To achieve an approximate bijective solution, one remedy is to use an auxillary network to represent the backward warp; however this will introduce computational overhead. To this end, we propose explicitly modeling inversions in the neural network architectures, formally learning an Invertible Neural Network (INN). Our results demonstrate that opting for an architecture that is explicitly invertible is more effective for jointly optimizing both the pose and radiance field, outperforming existing strong baselines~\cite{lin2021barf,chen2023local}. Notably, our INN-based approach achieves an improvement of over $50\%$ in pose accuracy when compared to the standard SE3 parameterization~\cite{lin2021barf}.

\section{Related Works}

\subsection{Joint NeRF and pose estimation}
Despite NeRF demonstrating compelling results in novel view synthesis, NeRF requires accurate camera poses. 
The differentiable nature of volume rendering used in NeRF facilitates the backpropagation through the scene representation to update the camera poses. NeRFmm~\cite{wang2021nerf} demonstrates the possibility of optimizing camera poses within the NeRF framework. BARF~\cite{lin2021barf} introduces a coarse-to-fine positional encoding scheduling to improve the joint optimization of NeRF and camera poses, and remains a widely adopted method. GARF~\cite{chng2022gaussian} and SiNeRF~\cite{xia2022sinerf} advocate for leveraging the smoothness inherent in non-traditional activations to mitigate the noisy gradients due to high frequencies in positional embeddings. In contrast, NoPe-NeRF~\cite{bian2023nope} uses monocular depth prior as a geometry prior to constrain the relative poses. 
SPARF~\cite{truong2023sparf} and SCNeRF~\cite{jeong2021self} demonstrate that using keypoint matches or dense correspondence can  constrain the relative pose estimates with ray-to-ray correspondence losses. Park~\etal~\cite{park2023camp} proposes a preconditioning strategy to enhance camera pose optimization. DBARF~\cite{chen2023dbarf}  proposes using low-frequency feature maps to address the joint optimization problem for generalizable NeRF. Bian \etal~\cite{bian2023porf} proposes a pose residual field which learn the pose corrections to refine the initial camera pose for neural surface reconstruction. Closest to our approach is a very recent work L2G by Chen~\etal~\cite{chen2023local}, which tackles the camera pose representation using an overparameterization strategy. However, we achieve overparameterization in different manner. Unlike L2G which learn an MLP to predict rigid $SE(3)$ transformations, we propose using an MLP to model the rigid warp function between the pixel and the ray space. Our work argues that while overparameterization can be achieved in different manner, it is tightly coupled with the regularization and constraints imposed. In our case, invertibility of warps becomes an essential constraint. 
\vspace{-1em}
\subsection{Overparameterization}
Overparameterization in deep learning involves employing models with a substantially greater number of parameters than the quantity of training data. Recent research \cite{li2018learning, srivastava2014dropout, neyshabur2015norm} demonstrates that neural networks, when overparameterized, can effectively generalize to new, unseen data, noting that an increase in parameters often correlates with a decrease in test error. Moreover, \cite{neyshabur2015norm} revealed that this ability to generalize does not necessarily require explicit regularization, suggesting that the optimization process of overparameterized neural networks inherently prefers solutions that are more likely to generalize well. Additionally, \cite{nguyen2020global, allen2019learning, arora2019fine, oymak2020toward} illustrates that overparameterized networks are capable of consistently finding a global minimum through gradient-based optimization methods. These insights highlight the remarkable ability of overparameterized neural networks to deliver accurate predictions on unseen data, emphasizing their robustness and efficacy in generalizing beyond the training dataset.
\vspace{-1em}
\subsection{Invertible Neural Networks for Deformation Fields}
Recent advancements in Invertible Neural Networks (INNs)~\cite{dinh2016density,behrmann2019invertible,chen2018neural} have broadened their use in the field of 3D deformation. These networks are particularly useful in modeling homermorphic deformation, where the mapping between any frames are bijective and continuous. Because of this capability, they are now being used in various areas, particularly modeling deformation in spatial~\cite{yang2021geometry,paschalidou2021neural,jiang2020shapeflow} and temporal~\cite{wang2023tracking, lei2022cadex,niemeyer2019occupancy,cai2022neural}. 

\section{Methodology}
We define the mathematical notations for the camera operations and the joint camera pose estimation in~\cref{subsec:prelim}. Further, we outline our approach in \cref{subsec:proposed}.
\vspace{-2em}
\subsection{Bundle-Adjust NeRF Preliminaries}\label{subsec:prelim}
\subsubsection{Camera pose}\label{subsubsec:explicit_notation}
We consider a set of $T$ input images as ${\{\mathcal{I}_t\}}_{t=1}^{T}$ taken by camera with intrinsic matrix $K \in \mathbb{R}^{3 \times 3}$. For each camera $t$ corresponding to these images, BARF-style approaches~\cite{lin2021barf} define its \textit{camera-to-world (C2W)} as 
$P = (\mathbf{R}_t, \mathbf{t}_t) \in SE(3)$, where $\mathbf{R}_t \in SO(3)$ and $\mathbf{t}_t \in \mathbb{R}^3$, respectively. 
\vspace{-1em}
\subsubsection{Camera projection}
For any vector $\mathbf{x} \in \mathbb{R}^{l}$ of dimension $l$, we define its homogeneous representation $\mathbf{\bar x} \in \mathbb{R}^{l+1}$ as $\mathbf{\bar x} = [\mathbf{x}^T, 1 ]$. We define $\pi$ as the camera projection operator, which maps a 3D point in the \textit{camera coordinate frame} denoted as $\mathbf{x}^{(C)} \in \mathbb{R}^3$ to a corresponding 2D pixel coordinate $\mathbf{u} \in \mathbb{R}^2$. $\pi^{-1}$ denotes the camera backprojection that maps a pixel $\mathbf{u}$ coordinate and depth $z$ to a 3D point in the camera coordinate as $\mathbf{x}^{(C)}$, for e.g., $\pi(\mathbf{x}^{(C)}) \cong K\mathbf{x}^{(C)}$ and $\pi^{-1}(\mathbf{u}, z) = z K^{-1}\mathbf{\bar u}$. We use $^{(C)}$ and $^{(W)}$ to denote that it is defined within the camera and world coordinate system respectively. 
\vspace{-1em}
\subsubsection{NeRF}
NeRF represents the volumetric field of a 3D scene as $f(\gamma(\mathbf{x}), \gamma(\mathbf{d})) \rightarrow (\mathbf{c},\sigma)$, which maps a 3D location $\mathbf{x} \in \mathbb{R}^3$ and a viewing direction $\mathbf{d}$ to a RGB color $\mathbf{c} \in \mathbb{R}^3$ and volume density $\sigma \in \mathbb{R}$. $\gamma: \mathbb{R}^3 \rightarrow \mathbb{R}^{3+6L}$ is the positional embedding function with $L$ frequency bases~\cite{mildenhall2021nerf}. This function is parameterized using an MLP as $f_{\mathbf{\Theta}_{rgb}}$.
Given $T$ input images $\{\mathcal{I}_t\}_{t=1}^{T}$ with corresponding camera poses $\{P_{t}\}_{t=1}^{T}$, NeRF is optimized by minimizing photometric loss $\mathcal{L}_{rgb}$ between synthesized images $\mathcal{\hat I}$ and original image $\mathcal{I}$ as 
\begin{equation}\label{eq:nerf_objfn}
    \min_{\mathbf{\Theta}_{rgb}} \,\,{\sum_{t=1}^{T} \sum_{\mathbf{u} \in \mathbb{R}^{2}}\| \mathcal{\hat{I}}(\mathbf{u}, {P}_{t}; \mathbf{\Theta}_{rgb}) - \mathcal{I}_{i}(\mathbf{u}) \|_{2}^{2}},
\end{equation}
\subsubsection{Volume rendering}
For simplicity, let's start by assuming the rendering operation of NeRF operates in the \textit{camera coordinate system}. We will generalize this later. Each pixel coordinate determines a viewing direction $\mathbf{d}$ in the camera coordinate system, whose origin is the camera center of projection $\mathbf{o}^{(C)}$. We can define a 3D point along the camera ray associated with $\mathbf{{u}}$ sampled at depth $z_i$ as $\mathbf{r}^{(C)}(z) = \mathbf{o}^{(C)} + z_{i,u}K^{-1}\mathbf{\bar{u}}$.~\footnote{This can be succinctly written as  $\mathbf{r}^{(C)}(z) = z_{i,u} \mathbf{d}$ as $\mathbf{o}^{(C)}$ is $[0,0,0]^{T}$ in camera coordinate space.} To render the colour of $\mathcal{\hat I}_{i,u}$ at pixel coordinate $\mathbf{u}$, we sample $M$ discrete depth values along the ray between a near bound $z_n$ and a far bound $z_f$. For each sampled value, we query the NeRF $f_{{\Theta}_{rgb}}$ to obtain the corresponding radiance fields. The output from NeRF is then aggregated to render the RGB colour as
\begin{equation}\label{eq:volume_rendering}
    \hat{\mathcal{I}}(\mathbf{u}) = \int_{z_{n}}^{z_{f}} T(\mathbf{u}, z) \sigma(\mathbf{r}(z)) \mathbf{c}(\mathbf{r}(z)) \delta z,
\end{equation}
where $T(\mathbf{u}, z) = \exp(-\int_{z_{n}}^{z} \sigma(\mathbf{r}(z))) \delta z'$ denotes the accumulated transmittance value along the ray. We refer readers to~\cite{mildenhall2021nerf, levoy1990efficient} for more details of volume rendering operation. In practice, \cref{eq:volume_rendering} is approximated using $M$ points sampled along the ray at depth $\{{z}_{i}\}_{{i=1}}^M$, which produces radiance field outputs as $\{\mathbf{y}_{i}\}_{{i=1}}^M$. Denoting the ray compositing function in \cref{eq:volume_rendering} as $g(.) \in \mathbb{R}^{4M} \rightarrow \mathbb{R}^{3}$, we can rewrite $\tilde{\mathcal{I}}(\mathbf{u}) = g( \{\mathbf{y}_i\}_{i=1}^{M})$. Finally, given a camera pose $P$, we can then transform the ray $\mathbf{r}^{(C)}$ to the \textit{world coordinate} to obtain $\mathbf{r}^{(W)}$ through a 3D rigid transformation $\mathcal{T}$. The rendered image is then obtained as
\begin{equation}
    \mathcal{\hat{I}}(\mathbf{u},\mathbf{p}) = g\bigg( \bigg \{ f\big(\mathcal{T}( \mathbf{r}^{(C)}(z),P);\mathbf{\Theta}_{rgb}\big) \bigg \} _{i=1}^{M}\bigg). 
\end{equation}

\subsubsection{Joint optimization of pose and NeRF}\label{subsec:method_barf}
Prior works~\cite{lin2021barf, wang2021nerf, chng2022gaussian,xia2022sinerf} demonstrate that it is feasible to optimize both camera pose and NeRF by minimizing $\mathcal{L}_{rgb}$ in \cref{eq:nerf_objfn}. This is achieved by considering $P$ as optimizable parameters, see \cref{subsubsec:explicit_notation} for its parameterization. Consequently, ray $\mathbf{r}$ is now dependent on the camera pose. Mathematically, this joint optimization can be rewritten as
\begin{equation}\label{eq:barf_objfn}
    \min_{P, \mathbf{\Theta}_{rgb}} \,\,{\sum_{t=1}^{T} \sum_{\mathbf{u} \in \mathbb{R}^{2}}\| \mathcal{\hat{I}}( \underbrace{\mathcal{T}( \mathbf{r}^{(C)}(z,\mathbf{\bar u}); {P}}_{G(.)}); \mathbf{\Theta}_{rgb}) - \mathcal{I}_{i}(\mathbf{u}) \|_{2}^{2}}.
\end{equation}
\begin{figure}[t]
    \centering
    \includegraphics[width=\textwidth]{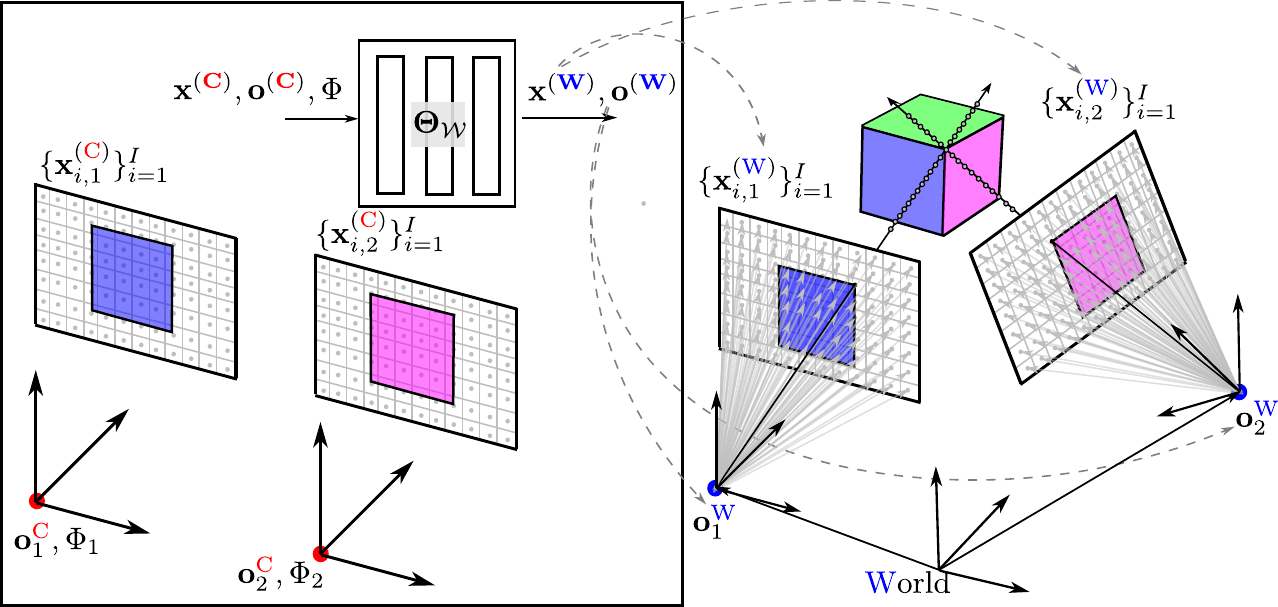}
    \caption{An overview of our INN-based approach, illustrated using two views $\mathcal{I}_1$ and $\mathcal{I}_2$. INN which is denoted as $h_{\mathbf{\Theta}_{\mathcal{W}}}$ takes the pixel locations in the camera coordinate system $\mathbf{x}_{i,t}^{(\textcolor{red}{C})}$, along with the frame-dependent latent code $\Phi_{t}$, and output its corresponding location in the world coordinate system as $\mathbf{x}_{i,t}^{(\textcolor{blue}{W})}$, see \cref{subsec:proposed} for full details. }
    \vspace{-1em}
    \label{fig:method}
\end{figure}
\vspace{-2em}
\subsection{Invertible Neural Warp for Ray Transform}\label{subsec:proposed}
We propose to overparameterize $P$ using an Invertible Neural Network (INN). There exist two options for parameterization: (i) 
use a separate INN for each camera $P_t$; (ii) use a single INN that is shared across all frames, coupled with a learnable code that is unique to frame $t$. Drawing inspiration from the dynamic NeRF method used for representing deformation fields~\cite{park2021nerfies,wang2023tracking,wang2023flow,cai2022neural} and also considering parameter efficiency, we have chosen to pursue the latter strategy for our proposed pose overparameterization -- a single, globally shared neural $\mathbf{\Theta}_{\mathcal{W}}$ network across all frames, coupled with an optimizable per-frame latent code $\Phi_t \in \mathbb{R}^{D}$, see supp. (Sec. C) for the comparison of using multiple-INNs versus single-INN. 
Consequently, we can rewrite $G(.)$ in \cref{eq:barf_objfn} as  $h( \mathbf{r}^{(C)}; {\mathbf{\Theta}_{\mathcal{W}}, \Phi_t})$, where $h(.) :\mathbb{R}^{3+D} \rightarrow \mathbb{R}^{3}$. \cref{fig:method} presents our approach. In our approach, we model each pixel in the camera coordinate system $\mathbf{x}_{i,t}^{(C)}$ as an individual ray. Our proposed INN is designed for transforming these rays from camera coordinate to the world coordinate. Specifically, our proposed INN takes in the pixel coordinates $\mathbf{x}_{i,t}^{(C)} $ and camera center $\mathbf{o}_t^{(C)}$, both defined in the camera coordinates $t$, coupled with the frame-dependent latent code and outputs their corresponding equivalent $\mathbf{x}_{i,t}^{(W)} $ and camera center $\mathbf{o}_t^{(W)}$ in the world coordinates. 

\subsubsection{Rigidity prior}
\vspace{-1em}
In our formulation, each pixel is represented as an individual ray within the camera coordinate system. This inherently relaxes the rigidity constraint. As a result, the output from the INN does not necessarily conform to a global rigid motion. Given known camera-world correspondences ($\mathbf{x}_{i,t}^{(C)}, \mathbf{x}_{i,t}^{(W)}$), we can solve a closed-form rigid registration problem to determine a global pose, which can be integrated into our optimization problem as a rigidity prior $\mathcal{L}_{rigid}$ 
\vspace{-1em}
\begin{align}\label{eq:rigidity_prior}
     \min_{T^{*}} \sum_{i=1}^{L} \| \mathbf{x}_{i,t}^{(C)} - T^{*} \circ \mathbf{x}_{i,t}^{(W)} \|_{2}^{2}.
\end{align}
\subsubsection{Final optimization problem}
\vspace{-1em}
We solve our final optimization problem as
\begin{equation}\label{eq:barf_objfn}
    \min_{ \Phi_t, \mathbf{\Theta}_{\mathcal{W}}, \mathbf{\Theta}_{rgb}} \,\,{\sum_{t=1}^{T} \sum_{\mathbf{u} \in \mathbb{R}^{2}}\| \mathcal{\hat{I}}( h( \mathbf{r}^{(c)}; {\mathbf{\Theta}_{\mathcal{W}}, \Phi_t}) ; \mathbf{\Theta}_{rgb}) - \mathcal{I}_{i}(\mathbf{u}) \|_{2}^{2}} + \lambda \mathcal{L}_{rigid}.
\end{equation}

\subsection{Advantages of INN for Overparameterizing Rigid Ray Warps}
BARF-based approach parameterizes camera pose $P$ of each frame using a $SE(3)$ (see \cref{subsubsec:explicit_notation}), which guarantees that $\mathcal{T}$ is a bijective mapping. Therefore, when overparameterizing camera poses, it is crucial that the neural network adheres to the bijection property because this one-to-one correspondence ensures that there is a unique output in the \textit{world} space for every point in the \textit{camera} space.
As we will demonstrate in \cref{subsec:exp_2d_result}, simply applying a rudimentary strategy (denoted as Naive) when overparameterizing the rigid warps of ray (camera-world) with an MLP often does not achieve convergence, see \cref{fig:planar_exp_mlp} and \cref{tab:planar_exp}. Consequently, to attain this bijective property using an MLP, it is necessary to introduce an auxillary network to model the backward warps (Implicit-Invertible MLP). While effective, it presents a significant drawback: it results in a twofold rise in the computational complexity due to the existence of the backward network to enforce the self-consistency.
To mitigate this substantial increase in computational demands inherent in the modified MLP approach, we propose the use of INNs for parameterize these bijections. INN implements the bijective mappings by composing affine transformations into several blocks. Within each block, the input coordinates are divided into two segments; the first part remains constant and is used to parameterize the transformation that is applied to the second part~\cite{cai2022neural}. Besides their inherent invertiblity, INN also offers the advantage of homeomorphic property, which potentially facilitate a more flexible optimization trajectory that is less susceptible to a suboptimal minimum trajectory, see \cref{subsec:homeo}

\section{Experiments}
\subsection{Baselines}\label{subsec:exp_baselines}
We compare our approach with two representative methods in pose-NeRF joint optimization: the standard global $SE3$-approach BARF~\cite{lin2021barf}, and the overparameterized representation L2G~\cite{chen2023local}. For all experiments, we use the original implementations including their default settings for coarse-to-fine scheduling, architecture and hyperparameters, see supp. (Sec. A) for more details. Additionally, in our 2D planar experiments, we include a comparison with another two variants called the Naive (MLP) and Implicit-Invertible MLP to execute ray transform. This comparison is specifically designed to highlight the significance of invertibility when employing MLPs for executing ray transformation. 
\vspace{-1em}
\subsubsection{Local-to-global (L2G)~\cite{chen2023local}}
L2G uses an MLP to predict rigid $SE3$ transformation for each ray. These predicted transformation parameters are then used to analytically estimate the transformed coordinates in the world space.
\vspace{-1em}
\subsubsection{Naive}\label{subsubsec:mlp}
This is the simplest version of our baseline that uses one primary network, denoted as $h_{fwd} :(\mathbf{x}^{(C)}, {\Phi}_t) \rightarrow \mathbf{x}^{(W)}$, to learn the forward mapping which takes in the coordinates from the camera space (coupled with a per-frame latent code), and outputs the corresponding coordinates in the world space.
\vspace{-1em}
\subsubsection{Implicit-Invertible MLP}\label{subsubsec:iimlp}
We use two networks $h_{fwd}$ and $h_{bwd}$ to enforce approximate invertibility. Alongside primary network $h_{fwd}$, we use a secondary network $h_{bwd} :(\mathbf{x}^{(W)}, {\Phi}_t) \rightarrow \mathbf{\hat x}^{(C)}$ to invert the outputs from the primary network $h_{fwd}$. To minimize deviations from bijections, we introduce a regularization term $\mathcal{L}_{implicit}$ as $\| \mathbf{x}^{(C)} - \mathbf{\hat x}^{(C)}\|_{2}^{2}$ into the optimization problem \cref{eq:barf_objfn}.


\vspace{-1em}
\subsubsection{Explicit-Invertible INN (Ours)}\label{subsubsec:eimlp}
We explicitly model inversions in the neural network architecture by formally learning an INN. We have chosen to utilize architecture proposed by NDR-INN~\cite{cai2022neural}, see supp. (Sec. A.1) for the architecture details that we use for all our experiments. 

\subsection{2D Planar Neural Image Alignment}\label{subsec:exp_2d}
Following BARF~\cite{lin2021barf, chen2023local, chng2022gaussian}, we learn a 2D neural image field, for creating a homography-based panoramic image from $N$ patches cropped from the original image, each generated with random homography perturbations. Specifically, we learn a 2D coordinate network $f(\mathbf{\Theta}_{rgb})$ to render the stitched image. Each pixel in the $N$ training patch is warped using the estimated homography $H$ to create the rendered image. 
We choose the ``cat'' image from ImageNet~\cite{deng2009imagenet}.
We initialize patch warps as identity and fix the gauge freedom by anchoring the first warp to align the neural image to the original image~\cite{lin2021barf, chng2022gaussian, chen2023local}. We randomly generate $20$ different homography instances, with scale-noise parameter $0.1$ and $0.2$ for homography and translation, respectively. We solve \cref{eq:rigidity_prior} for the homography using a Direct Linear Transform (DLT) solver~\footnote{\url{https://github.com/kornia/kornia}}.
\vspace{-1em}
\subsubsection{Experiment settings.} We evaluate our proposed method against BARF~\cite{lin2021barf}, and our three overparameterized network variants: Naive, Implicit- and Explicit-Invertible MLPs, as detailed in~\cref{subsec:exp_baselines}. For both naive and Implicit-Invertible MLPs, we utilized a Leaky-ReLU MLP with five $256$-dimensional hidden units, and a $16$-dimensional latent code to represent the frame-dependent embeddings $\phi_t$. We also follow the default coarse-to-fine scheduling established by BARF. In the case of Implicit-Invertible MLP, we use the same architecture both $h_{fwd}$ and $h_{bwd}$. 
We use the Adam optimizer~\cite{kingma2014adam} to optimize for both the network weights $\mathbf{\Theta}_{rgb}$ and $\mathbf{\Theta}_{\mathcal{W}}$. We set the learning rate for both $\mathbf{\Theta}_{rgb}$ and $\mathbf{\Theta}_{\mathcal{W}}$ at $1 \times 10^{-3}$, with both decaying exponentially to $1 \times 10^{-4}$ and $1 \times 10^{-5}$, respectively. We set the weighting term for $\mathcal{L}_{rigid}$ for both overparameterized MLPs to $1 \times 10^{2}$ and the consistency term $\mathcal{L}_{implicit}$ for Implicit-Invertible MLP to $1 \times 10^{1}$.
\vspace{-1em}
\begin{figure}[t]
    \centering
    \includegraphics[width=\textwidth]{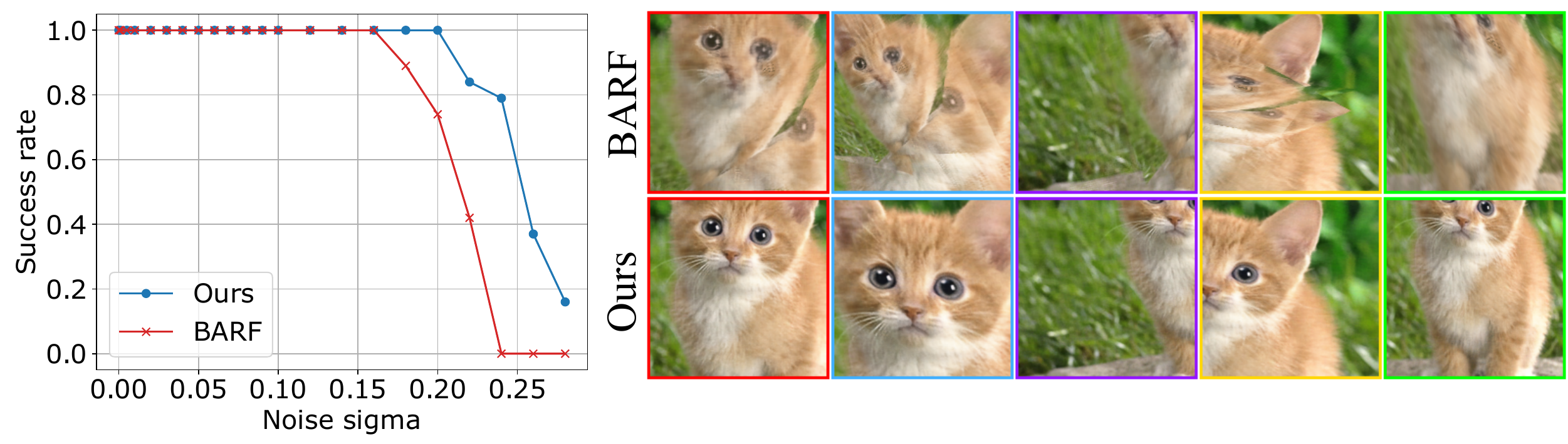}
    \vspace{-2em}
    \caption{Basin convergence analysis of our approach versus BARF in a 2D planar experiment. On the left, we show the success rate of 20 runs, where we initialize the homography using \textit{groundtruth}, and gradually introduced noise perturbations to the translation component. The noise scale is varied from $0$ to $0.30$. We used 5-pixel threshold to determine success convergence. Notably, our approach (blue) demonstrates higher noise tolerance compared to BARF (red). On the right, we show a qualitative comparison when both methods are perturbed with the highest magnitude of noise.}
    \vspace{-2em}
    \label{fig:exp_homo_analysis}
\end{figure}

\subsubsection{Robustness to noise perturbations}
~\cref{fig:exp_homo_analysis} analyzes the robustness of BARF and our approach under different noise perturbations across $20$ different homography instances. Each run began with the initialization of the homography using the ``groundtruth'', and noise perturbations were gradually introduced, ranging from $0$ to $0.3$ to the translation component of the homography. ~\cref{fig:exp_homo_analysis} indicates that our representation exhibits a higher tolerance to noise compared to BARF.
\vspace{-1em}
\subsubsection{Results.}\label{subsec:exp_2d_result}
~\cref{tab:planar_exp} summarizes the statistical results from $20$ runs, where we report the warp error and patch reconstruction error. We quantify the warp error in terms of corner error, defined as the L2 distance between the groundtruth corner position and estimated corner position, and PSNR as the metric to assess the reconstruction quality. We used $5$-pixel threshold to define the success convergence.
\cref{fig:planar_exp} presents a qualitative result for a homography instance. By enforcing approximate invertibility, the Implicit-Invertible MLP demonstrates a significantly higher rate of successful convergence compared to the naive version of MLP, with success rates improved by $65\%$. We further show that by \textit{explicitly} injecting invertibility into the architecture (Ours), the success rate is increased to $75\%$. 
This result reinforces that \textit{invertibility} is crucial when learning rigid warp functions via overparameterization, and using an architecture that guarantees bijective property is effective in ensuring the pose converge to an optimal solution during joint optimization in practice. Henceforth, we will focus exclusively on using our INN-based approach for subsequent results. For completeness, we also compare with another setup, where each frame is parameterized with a neural network, see supp. (Sec. B.1). Interestingly, apart from the parameter efficiency, we find that using a single global neural network for all frames is sufficient to converge to a good pose solutions. We hypothesized that the difference may be attributed to the benefits of gradient sharing in the shared neural network setup. This result is aligned with the findings by Bian \etal~\cite{bian2023porf}. Consequently, we adhere to the design where we use one single INN shared across all the frames, coupled with a frame-specific latent code for the rest of our experiments.

\begin{table}[t]
\centering
\caption{Statistical result for 20 homography runs, with scale noise of $0.1$ for homography and $0.2$ for translation. The warp error is quantified in terms of corner error, and the patch reconstruction error in measured in PSNR. We provide mean and std dev for the evaluation. We used 5-pixel threshold to determine success convergence. }
\vspace{-1em}
\label{tab:planar_exp}
\setlength\tabcolsep{6pt} 
\resizebox{\columnwidth}{!}
{
\begin{tabular}{l|cc|cc|c}
\toprule
& \multicolumn{2}{|c|}{Corner error (px) $\downarrow$} & \multicolumn{2}{|c|}{Patch PSNR $\uparrow$} 
& \multicolumn{1}{|c}{Success rate $\uparrow$} \\
& \multicolumn{1}{|c}{Mean} & \multicolumn{1}{c|}{Std. dev.} & \multicolumn{1}{|c}{Mean} & \multicolumn{1}{c|}{Std. dev.} & \multicolumn{1}{|c}{ (Upper bound:1.00)}  \\
\midrule
BARF~\cite{lin2021barf}  & 29.63 & 28.18 &  28.94 & 4.38 & 0.30   \\
\midrule
Naive  & 85.59 & 30.31 &  25.86 & 2.07 & 0.00\\
Implicit-Invertible  & 13.92 & 22.93 &  33.70 & 3.93 & 0.65  \\
Explicit-Invertible (INN)  & \textbf{4.70} & \textbf{6.47} & \textbf{34.71} & \textbf{2.37} & \textbf{0.75} \\
\bottomrule
\end{tabular}
}
\end{table}

\subsection{Neural Radiance Fields (NeRF)}
In this section, we compare our proposed representation with BARF~\cite{lin2021barf} and L2G~\cite{chen2023local}. We assume known intrinsics for all methods. We perform our experiments on both the LLFF~\cite{mildenhall2019local} (\cref{subsec:exp_llff}), DTU~\cite{jensen2014large} (\cref{subsec:exp_dtu}) as well as Blender datasets in supp. (Sec. B2). We solve \cref{eq:rigidity_prior}  for the global rigid $SE3$ transformation using Umeyama algorithm~\footnote{\url{https://github.com/naver/roma}}.
\vspace{-1em}
\subsubsection{Evaluation metrics.}\label{subsubsec:evaluation_metric}
For pose estimation, we report the accuracy of the poses after globally aligning the optimized poses to the groundtruth~\cite{lin2021barf,chng2022gaussian,truong2023sparf,chen2023local}.~\footnote{For our proposed method, we evaluate the estimated global poses \cref{eq:rigidity_prior}.} We assess view synthesis using PSNR, SSIM and LPIPS. A standard procedure in view synthesis evaluation involves performing test-time photometric optimization on the trained models. This additional step is intended to \emph{factor out the pose errors}, which may otherwise compromise the quality of the synthesized views~\cite{lin2021barf, chng2022gaussian, truong2023sparf, chen2023local,tancik2023nerfstudio}. This process is akin to a \textit{pose refinement} method which minimises the photometric error on the synthesized image while keeping the trained NeRF model fixed. However, it is important to recognize that this pose correction may not accurately represent the initial accuracy of the methods in terms of pose estimation for view synthesis. Therefore, we opt to report the view synthesis quality both \textit{before} and \textit{after} the pose refinement step. 
On DTU, we extend our evaluation to include comparisons of rendered depth with ground-truth depth using mean depth absolute error, as well as reconstruction accuracy using Chamfer distance. For the Chamfer evaluation, we utilize the optimized poses estimated from all methods and employ a neural surface reconstruction algorithm called Voxurf~\cite{wu2022voxurf} for geometry reconstruction. Further details on pose alignment and metrics computations in the supp. (Sec. A). 
\begin{figure}[t]
    \centering
    \begin{subfigure}{0.32\textwidth}
        \includegraphics[height=3.5cm, width=\linewidth]{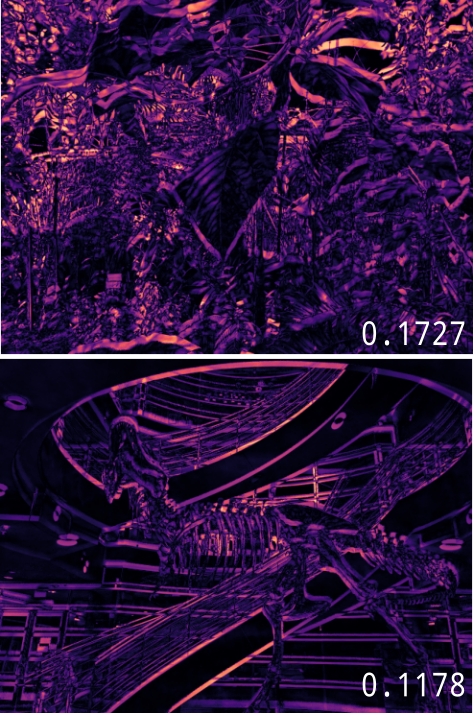}
        \caption{BARF~\cite{lin2021barf}}
        \label{fig:sub1}
    \end{subfigure}
    \begin{subfigure}{0.32\textwidth}
        \includegraphics[height=3.5cm, width=\linewidth]{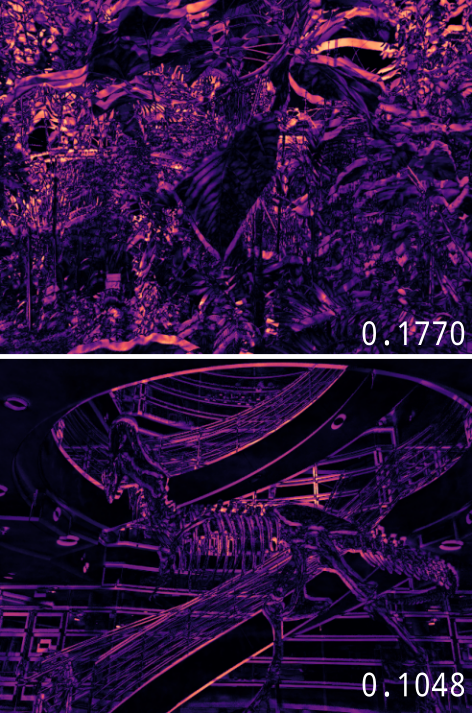}
        \caption{L2G~\cite{chen2023local}}
        \label{fig:sub2}
    \end{subfigure}
    \begin{subfigure}{0.32\textwidth}
        \includegraphics[height=3.5cm, width=\linewidth]{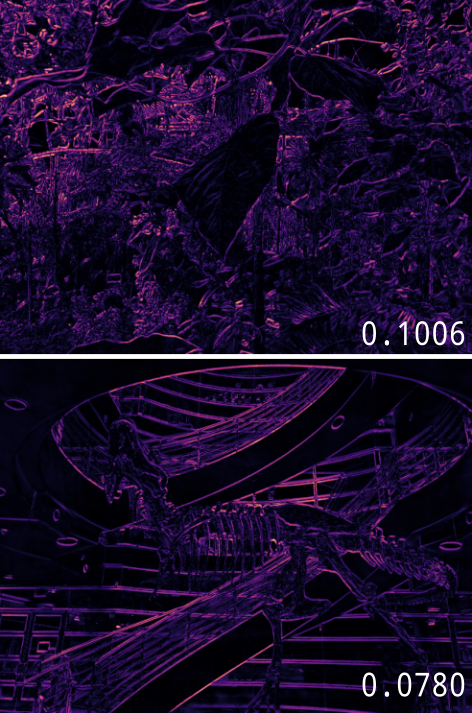}
        \caption{INN(Ours)}
        \label{fig:sub3}
    \end{subfigure}
    \vspace{-1em}
    \caption{Qualitative analysis of reconstruction error on leaves (\textbf{top}) and trex (\textbf{bottom}). We present the average image reconstruction error through insets. Our approach presents the lowest misalignment error, as indicated darker areas in the error map.}
    \label{fig:llff_error_map} 
\end{figure}

\subsection{Forward Facing Scenes: LLFF}\label{subsec:exp_llff}
\subsubsection{Experiment settings.}The standard LLFF benchmark dataset~\cite{mildenhall2019local} consists of eight real-world, forward-facing scenes captured using hand-held cameras. Following~\cite{lin2021barf,truong2023sparf, chng2022gaussian}, we initialize all camera poses with the \textit{identity} transformations for all the methods. We employ the same training and testing split as in BARF~\cite{lin2021barf}. We use the evaluation metrics described in \cref{subsubsec:evaluation_metric}. 
\vspace{-1em}
\subsubsection{Implementation details.} We train all methods for $200$k iterations and randomly sample $2048$ pixel rays at each optimization step~\cite{lin2021barf,chng2022gaussian,chen2023local}. We train without hierarchical sampling. We set the learning for  $\mathbf{\Theta}_{rgb}$ to be $1 \times 10^{-3}$ decaying to $3 \times 10^{-4}$, and $5 \times 10^{-4}$ for $\mathbf{\Theta}_{\mathcal{W}}$ decaying to $1 \times 10^{-6}$. We also follow the default coarse-to-fine scheduling by BARF~\cite{lin2021barf}, see supp. for full details.
\vspace{-1em}
\subsubsection{Results.} Our approach achieves a substantial reduction of rotation errors ($70 \%$ vs. BARF and $35 \%$ vs. L2G) and translation errors ($50 \%$ vs. BARF and $20\%$ vs. L2G). Additionally, the superior performance of both our method and L2G over BARF further highlight the merits of overparameterization for simultaneous pose and neural field estimation. This significant improvement in camera pose accuracy directly enhances the performance of view synthesis, as evident in \cref{tab:llff}. In particular, instances such as \textit{trex} and \textit{leaves} observe a significant improvement. \cref{fig:llff_error_map} illustrates the absolute error between the original image and the rendered image on these two instances. Our approach presents the lowest misalignment error, as indicated darker areas in the error map. It is important to note, however, while L2G and our approach appear comparable after test-time optimization, this refinement procedure has mitigated the pose estimation noise to improve PSNR, as discussed in \cref{subsubsec:evaluation_metric}. For more qualitative results and ablation studies, please refer to supp. (Sec. B).

\begin{table}[t]
\centering
\vspace{-1em}
\caption{Evaluation of the LLFF dataset~\cite{mildenhall2019local} using initial \textit{identity} poses. We compute results for BARF~\cite{lin2021barf} and L2G~\cite{chen2023local} using their codebases. The results are \textbf{averaged} over all \textbf{eight} scenes, see supp. (Sec. B) for a breakdown of results per scene. }
\label{tab:llff}
\setlength\tabcolsep{6pt} 
\resizebox{\columnwidth}{!}{
\begin{tabular}{l|cc|ccccccc}
\toprule
& \multicolumn{2}{|c|}{Pose accuracy} & \multicolumn{6}{|c}{Novel view synthesis}  \\
& \multicolumn{1}{|c}{Rotation} & \multicolumn{1}{c|}{Translation} & \multicolumn{3}{|c}{\textit{Before} test-time} & \multicolumn{3}{c}{\textit{After} test-time} \\
&\multicolumn{1}{|c} ($\degree$) &  (\(\times 100\)) & \multicolumn{1}{|c}{PSNR}  & SSIM  & LPIPS  & PSNR & SSIM  & LPIPS  \\
\midrule
BARF~\cite{lin2021barf}  & 0.90 & 0.40 & 17.00  & 0.41 & 0.32  & 23.82 & 0.72 & 0.24 \\
L2G~\cite{chen2023local} & 0.48 & 0.30 & 17.99 & 0.46 & 0.26 & \textbf{24.35} & \textbf{0.75} & \textbf{0.21} \\
\midrule
INN(Ours) & \textbf{0.31} & \textbf{0.24} & \textbf{19.31} & \textbf{0.52} & \textbf{0.25} & 24.28 & 0.74 & 0.22  \\
\bottomrule
\end{tabular}
}
\end{table}
\vspace{-1em}
\subsection{Homeomorphism perspective: A qualitative analysis}\label{subsec:homeo}
\begin{figure}[t]
    \centering
    \includegraphics[width=\textwidth]{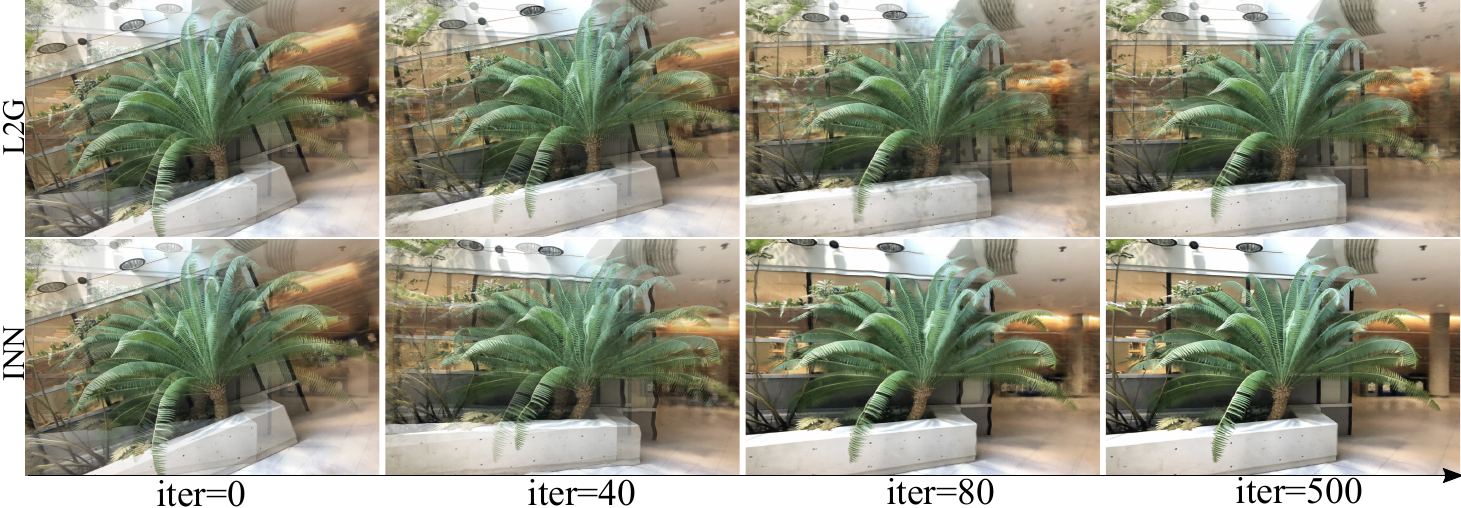}
    \vspace{-1em}
    \caption{Qualitative analysis of intermediate rendered image compared to superimposed Groundtruth (lighter visualization) when using L2G~\cite{chen2023local} and our approach on single-view pose estimation. Notably, when using our approach for pose estimation, we observe noticeable \textit{deformation} in the rendered scene depicted in the bottom row (zoom in for better view). These deformations indicate that at each iteration, the INN predicts general homeomorphisms that are not rigid transformation. Thus yielding a flexible optimization trajectory that does not land in a suboptimal minimum trajectory.}
    \label{fig:inerf_qualitative}
\end{figure}

We present a qualitative analysis on a single-view pose estimation that sheds light on the empirical effectiveness of our approach compared to the L2G method \cite{chen2023local}. Our method leverages the concept of an INN, which predict homeomorphisms—continuous, invertible transformations not limited to the rigid motions of the $SE(3)$ group. Unlike the L2G method, which constrains pose estimation within the rigid bounds of $SE(3)$, our INN-based approach embraces a broader spectrum of transformations. This grants the optimization process a higher degree of flexibility, allowing for a diverse range of optimization paths and facilitating a smoother trajectory towards the solution.
To empirically validate our hypothesis, we conducted experiments using a trained NeRF model to estimate the camera pose relative to a 3D scene, aiming to minimize the photometric error between NeRF-rendered and actual observed images. Despite starting from the same initial pose (off by $20^{\degree}$), and employing a random sampling of 2048 rays per iteration for all methods, our INN approach outperformed the L2G method. The L2G method often converged to suboptimal poses, whereas the INN method achieved accurate pose estimation, as evidenced in \cref{fig:inerf_qualitative} where the NeRF-rendered images is well-aligned with the groundtruth. This success can be attributed to the INN's ability to predict a general homeomorphim, a transformation more general than a rigid transformation, which offers a significant advantage in navigating the optimization landscape more effectively and avoiding suboptimal local minima.

\begin{table}[t]
\centering
\vspace{-1em}
\caption{Absolute pose accuracy evaluation of the DTU dataset~\cite{jensen2014large} using initial \textit{noisy} poses that corresponds to an average rotation and translation error of $15 \degree$ and 70 respectively. The \textit{upper} section shows rotation errors in degrees, while the \textit{lower} section displays translation errors ($\times 100$). {\colorbox{tabfirst}{red box}} denotes the \textbf{best} result.}
\vspace{-1em}
\label{tab:dtu_pose}
{\small 
\setlength\tabcolsep{3pt} 
\resizebox{\columnwidth}{!}{%
\begin{tabular}{ll|cccccccccccccc|c}
\toprule
\multicolumn{2}{c}{} & \multicolumn{15}{c}{Scan IDs | \textbf{Initial rotation err}: $\mathbf{15 \degree}$, \textbf{translation err ({$\times$}100)} \textbf{: 70} }\\
Methods & Metric & 24 & 37 & 40 & 55 & 63 & 65 & 69 & 83 & 97 & 105 & 106 & 110 & 114 & 118 &  \textbf{mean}\\
\midrule
BARF & &1.04 & 1.88 & \cellcolor{tabfirst}0.55  & 2.85  & 0.43  & 0.78 & 3.88 & \cellcolor{tabfirst}0.54 & 2.20 & 3.56 & 11.81  & \cellcolor{tabfirst}2.01 & 1.21 & 2.60  & 2.52 \\
L2G   & rotation & 0.90 & 3.07  & 0.57  & 10.81  & 0.74 & 0.94 & \cellcolor{tabfirst}1.55 & 0.59 & 0.64 & 2.85 & 13.15 & 5.92 & 2.90 & 12.47 & 4.08 \\
INN(Ours) & &\cellcolor{tabfirst}{0.21} & \cellcolor{tabfirst}0.48 & 0.73 & \cellcolor{tabfirst}1.64 & \cellcolor{tabfirst}0.40 & \cellcolor{tabfirst}0.62 & 2.75 & 0.55 & \cellcolor{tabfirst}0.54 & \cellcolor{tabfirst}0.30 & \cellcolor{tabfirst}3.89  & 2.91 & \cellcolor{tabfirst}1.06 & \cellcolor{tabfirst}0.25 & \cellcolor{tabfirst}1.17\\
\midrule
BARF & & 3.00 & 4.08 & 2.09  & 11.63  & 1.43  & 2.78  & 12.80  & 2.13  & 6.12  & 12.40 & 25.88 & \cellcolor{tabfirst}5.52 & 4.08 & 5.04 & 7.07 \\
L2G  & translation & 2.40 & 8.80  & 2.13  & 32.74 & 2.78 & 3.48  &  \cellcolor{tabfirst}5.08  & 2.31  & 2.37  & 5.34  & 36.20  & 14.01  & 8.64 & 37.16 & 11.67\\
INN(Ours) & & \cellcolor{tabfirst}0.50 & \cellcolor{tabfirst}1.01 & \cellcolor{tabfirst}0.97 & \cellcolor{tabfirst}5.24 & \cellcolor{tabfirst}0.96 & \cellcolor{tabfirst}0.78 & 5.30 & \cellcolor{tabfirst}0.67 & \cellcolor{tabfirst}1.19 & \cellcolor{tabfirst}0.43 & \cellcolor{tabfirst}9.96  & 11.52  & \cellcolor{tabfirst}3.78 & \cellcolor{tabfirst}0.64  & \cellcolor{tabfirst}3.07\\
\toprule
\end{tabular}
}
}
\end{table}

\begin{table}
\centering
\caption{Geometry evaluation of the DTU dataset~\cite{jensen2014large}, using initial \textit{noisy} poses that corresponds to an averange rotation and translation error of $15 \degree$ and 70 respectively. The \textit{upper} section presents mean absolute depth, while the \textit{lower} section displays reconstruction error measured by Chamfer distance. {\colorbox{tabfirst}{red box}} denotes the \textbf{best} result.}
\vspace{-1em}
\label{tab:dtu_geometry}
{\small 
\setlength\tabcolsep{3pt} 
\resizebox{\columnwidth}{!}{%
\begin{tabular}{llccccccccccccccc}
\toprule
\multicolumn{2}{c}{} & \multicolumn{15}{c}{Scan IDs | \textbf{Initial rotation err}: $\mathbf{15 \degree}$, \textbf{translation err ({$\times$}100)} \textbf{: 70} }\\
Pose & Metric & 24 & 37 & 40 & 55 & 63 & 65 & 69 & 83 & 97 & 105 & 106 & 110 & 114 & 118 &  \textbf{mean}\\
\midrule
BARF &  & 0.11 & 0.15 & \cellcolor{tabfirst}0.11  & \cellcolor{tabfirst}0.12  & \cellcolor{tabfirst}0.14  & 0.10 & 0.26 & \cellcolor{tabfirst}0.17 & 0.17 & 0.46 & 0.53  & 0.17  & 0.10 & 0.17 & 0.20\\
L2G  & depth $\downarrow$ & 0.08 & 0.24 & 0.12 & 0.38 & 0.17 & 0.12 & \cellcolor{tabfirst}0.11 & 0.18 & \cellcolor{tabfirst}0.10 &  0.18 & 0.24 & \cellcolor{tabfirst}0.17 & 0.12 & 0.84 & 0.22 \\
INN(Ours) & (abs) & \cellcolor{tabfirst}0.06 & \cellcolor{tabfirst}0.14 & 0.13 & \cellcolor{tabfirst}0.12 & 0.15 & \cellcolor{tabfirst}0.08 & 0.18 & \cellcolor{tabfirst}0.17 & 0.11 & \cellcolor{tabfirst}0.11 & \cellcolor{tabfirst}0.13 & 0.30 & \cellcolor{tabfirst}0.08 & \cellcolor{tabfirst}0.06 & \cellcolor{tabfirst}0.13 \\
\midrule
BARF & & 5.47 & 4.07 & \cellcolor{tabfirst}4.39 & 7.64 & 4.83 & 5.28 & 6.43 & 5.74 & 8.06 & 7.21 & 7.75 & 8.29 & 8.24 & 5.47 & 6.35 \\
L2G  & Chamfer $\downarrow$& 4.37 & 9.08 & 4.41 & 6.67 & 8.03 & 5.92 & \cellcolor{tabfirst}3.80 & 6.00 & 6.30 & 5.59 & \cellcolor{tabfirst}7.43 & \cellcolor{tabfirst}7.35 & 8.29 & 8.16 & 6.53 \\
INN(Ours) &  & \cellcolor{tabfirst}1.56 & \cellcolor{tabfirst}2.25 & 5.55 & \cellcolor{tabfirst}5.68 & \cellcolor{tabfirst}4.58 & \cellcolor{tabfirst}4.27 & 5.51 & \cellcolor{tabfirst}5.69 & \cellcolor{tabfirst}5.09 & \cellcolor{tabfirst}3.73 & 8.15 & 7.85 & \cellcolor{tabfirst}7.35  & \cellcolor{tabfirst}1.14 & \cellcolor{tabfirst}4.89 \\
\toprule
\end{tabular}
}
}
\end{table}
\vspace{-1em}
\subsection{360$\degree$ Scenes: DTU}\label{subsec:exp_dtu}
\subsubsection{Experimental settings.}
We evaluated on 14 test scenes from DTU~\cite{bian2023porf}. Following~\cite{lin2021barf,truong2023sparf, chen2023local}, we synthetically perturb the ground-truth camera poses with $15\%$ of additive Gaussian noise, which corresponds to an average rotation and translation error of $15\degree$ and $70$, respectively. For a fair comparison, we used the same initialization for all methods. We refer the readers to supp. (Sec B.5) for the results with Colmap initialization.
\vspace{-1em}
\subsubsection{Implementation details.}
As BARF~\cite{lin2021barf} and L2G~\cite{chen2023local} have not tested on DTU datasets, given that the DTU dataset encompass $360 \degree$ scenes which is similar to the original Blender dataset, we adopted the original hyperparameters used by the author for training BARF and L2G on the blender dataset. Following Bian \etal~\cite{bian2023porf}, we multiply the output of their local warp network by a small factor which is $\alpha=0.01$ for L2G~\cite{chen2023local}. 
For our approach, we set the learning rate for $\mathbf{\Theta}_{rgb}$ to be $1\times10^{-3}$ decaying to $1\times10^{-4}$, and  $\mathbf{\Theta}_{\mathcal{W}}$ to start from $5\times10^{-4}$ and decaying to $1\times10^{-8}$. We use the default coarse-to-fine scheduling by BARF~\cite{lin2021barf}.
\vspace{-2em}
\subsubsection{Results.}
\begin{figure}[h]
    \centering
    \begin{subfigure}[t]{1\textwidth}
        \includegraphics[width=\textwidth]{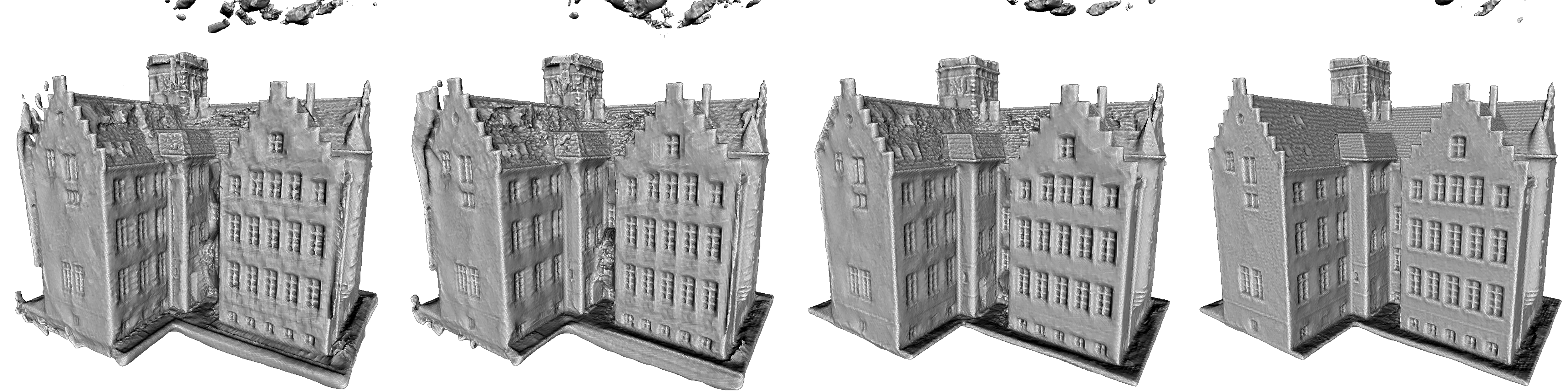}
        \label{fig:dtu_scan24}
    \end{subfigure}
    \begin{subfigure}[t]{1\textwidth}
        \includegraphics[width=\textwidth]{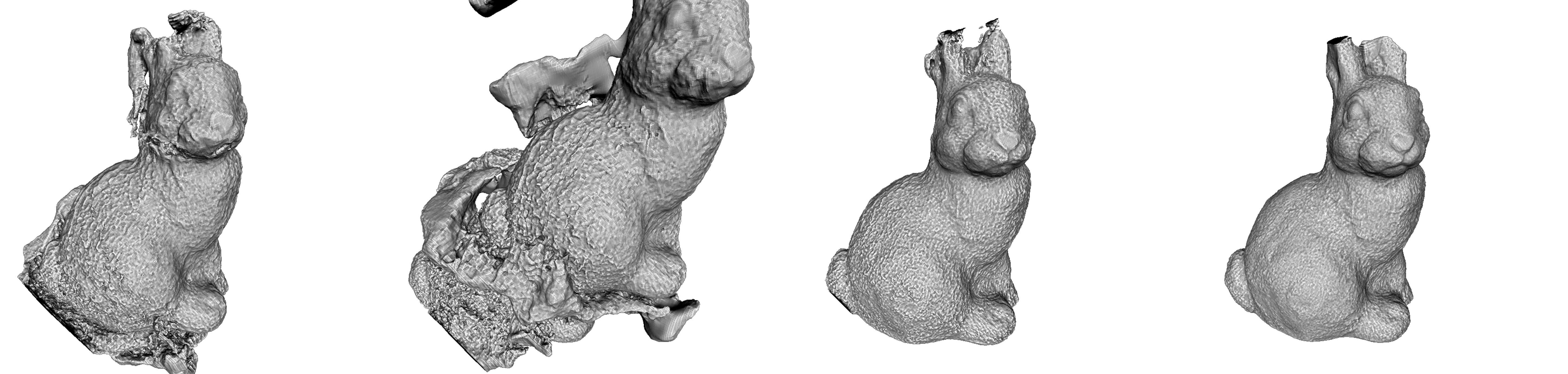}
        \label{fig:dtu_scan55}
    \end{subfigure}
    \begin{subfigure}[t]{1\textwidth}
        \includegraphics[width=\textwidth]{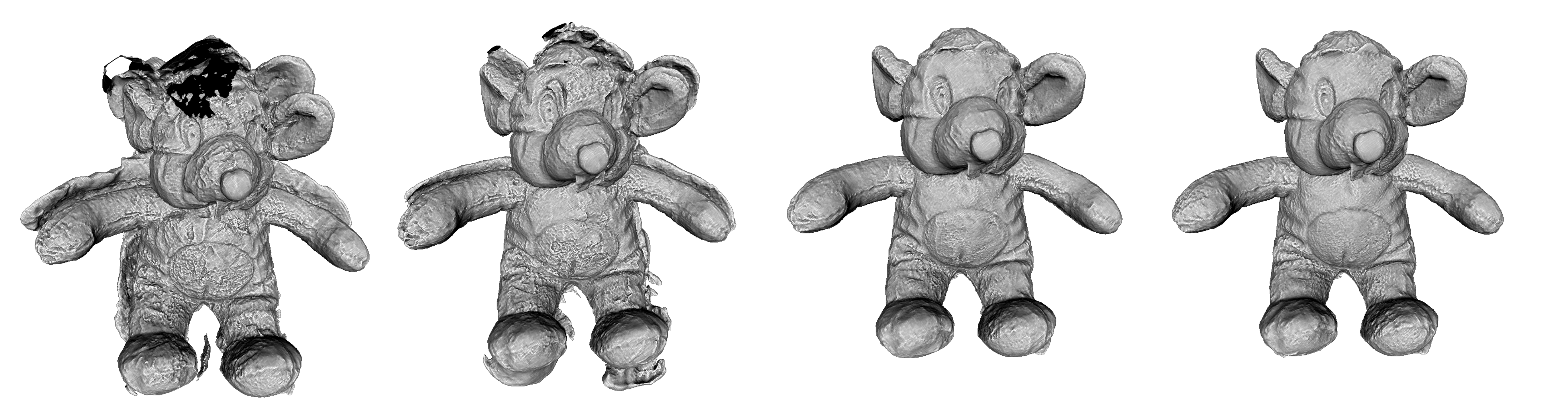}
        \label{dtu_scan105}
    \end{subfigure}
    \begin{subfigure}[t]{0.24\textwidth}
        \includegraphics[width=\textwidth]{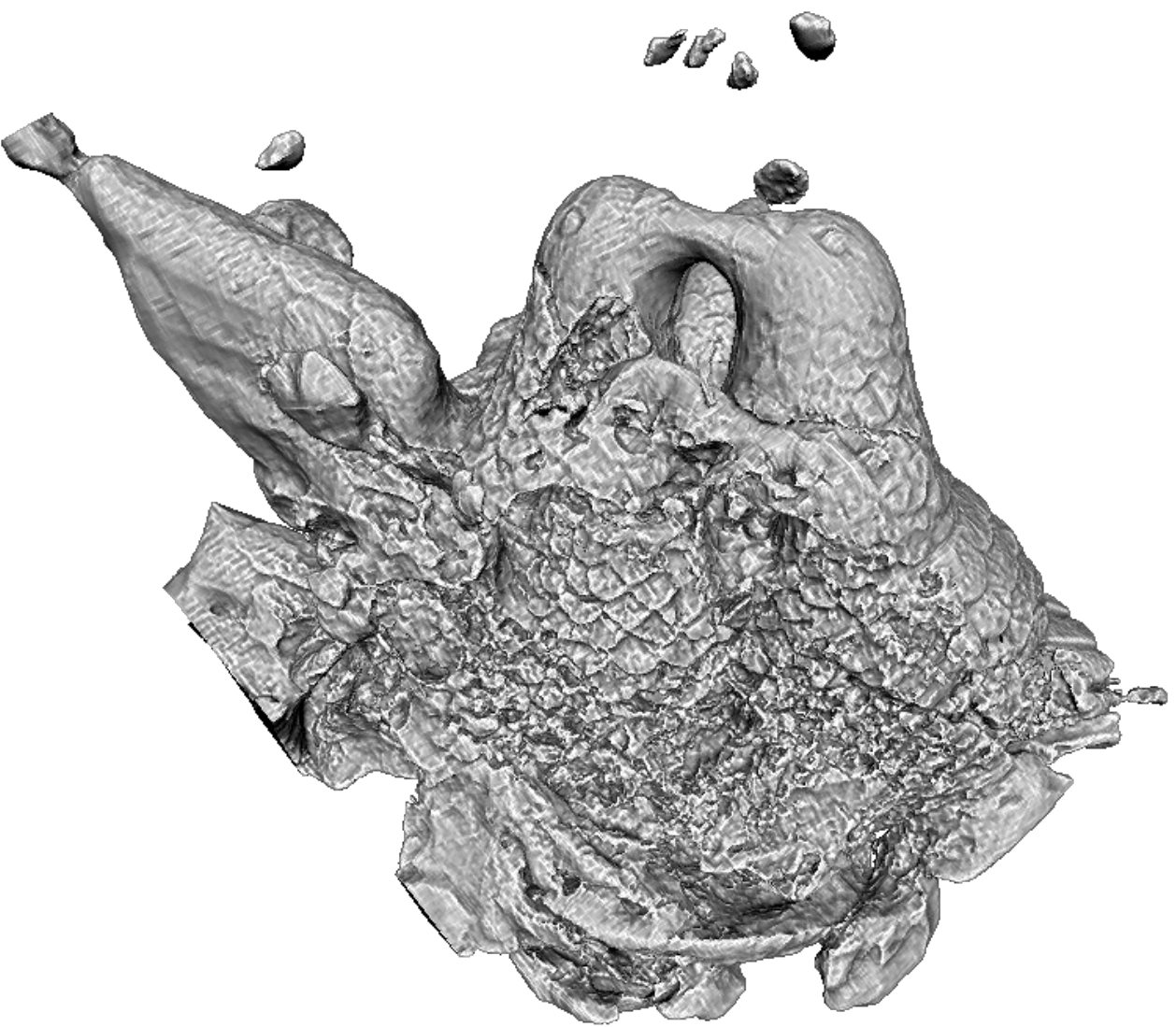}
        \caption{BARF~\cite{lin2021barf}}
        \label{dtu_scan105}
    \end{subfigure}
    \begin{subfigure}[t]{0.24\textwidth}
        \includegraphics[width=\textwidth]{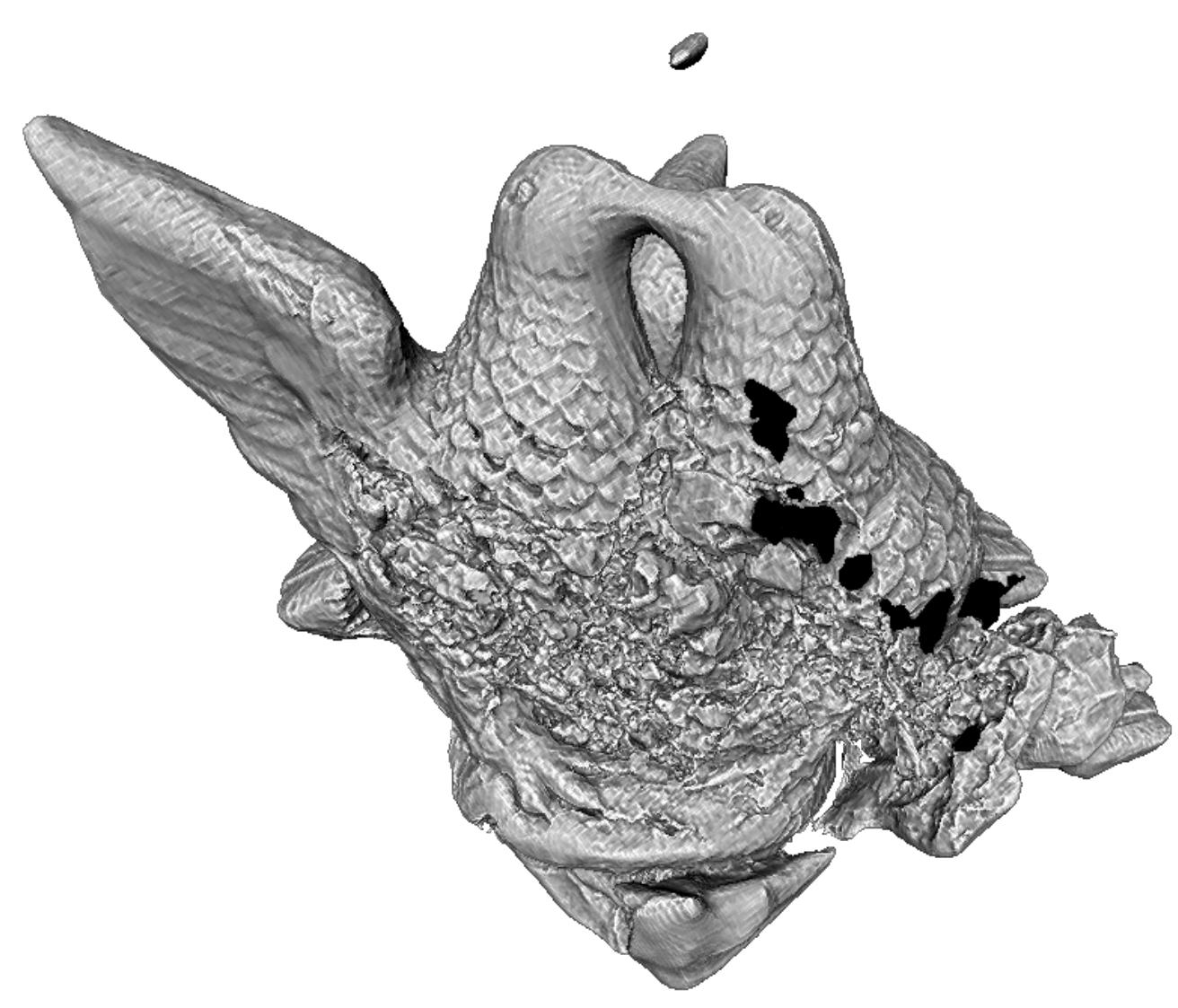}
        \caption{L2G~\cite{chen2023local}}
        \label{dtu_scan105}
    \end{subfigure}
    \begin{subfigure}[t]{0.24\textwidth}
        \includegraphics[width=\textwidth]{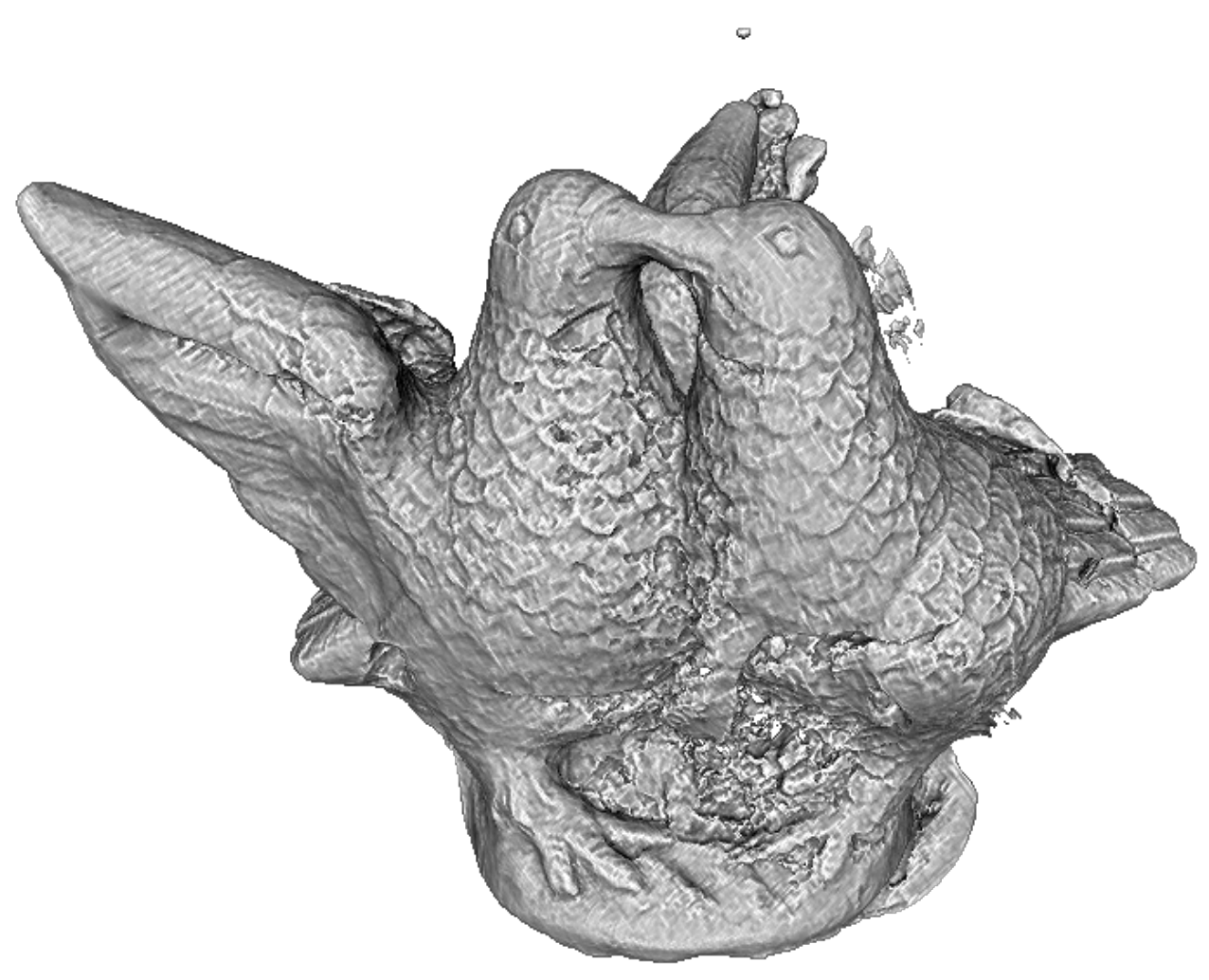}
        \caption{INN(Ours)}
        \label{dtu_scan105}
    \end{subfigure}
    \begin{subfigure}[t]{0.23\textwidth}
        \includegraphics[width=\textwidth]{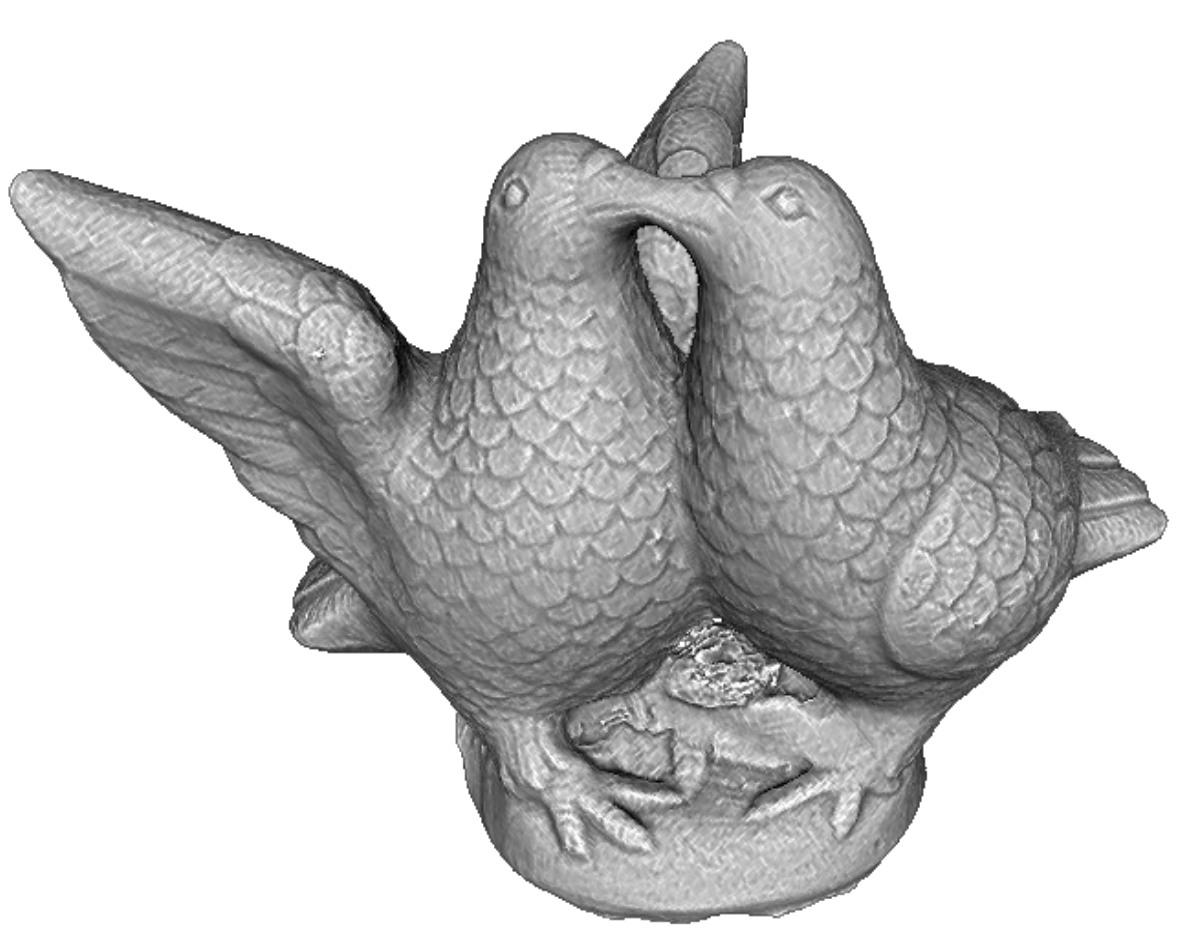}
        \caption{Groundtruth}
        \label{dtu_scan105}
    \end{subfigure}
    \caption{We showcase the reconstruction results for \textit{scan24}, \textit{scan55}, \textit{scan105} and \textit{scan106}. All meshes are generated using a neural surface reconstruction algorithm called Voxurf~\cite{wu2022voxurf} using the poses estimated by different approaches.}
    \label{fig:dtu_qualitative} 
\end{figure}

As demonstrated in \cref{tab:dtu_pose}, we outperform all baselines by a considerable margin across majority of the sequences in pose accuracy. Overall, our approach achieves approximately a $50\%$ improvement in rotation and a $60\%$ improvement in translation over BARF. When compared to L2G, our method shows a $70\%$ increase in accuracy for both rotation and translation. Additionally, our approach also consistently surpasses both baselines in geometry evaluation, as evidenced in the depth and reconstruction error in \cref{tab:dtu_geometry} and qualitative results in \cref{fig:dtu_qualitative}.  For the quantitative results for novel view synthesis and additional qualitative results, we refer the readers to see supp. (Sec B).

\section{Conclusion}
In this paper, we examine the benefits of overparameterizating poses via a MLP in the joint optimization task of camera pose and NeRF. We establish that invertibility is a crucial property. We further show that using an Invertible Neural Network, inherently equipped with a  guaranteed bijection property, significantly improves the convergence in pose optimization compared to existing representative methods.
\hfill \break

\clearpage
\hfill \break
\noindent \textbf{Acknowledgement} 
We thank Chee-Kheng (CK) Chng for insightful discussions and technical feedback.
%
%
\bibliographystyle{splncs04}
\bibliography{main}
\end{document}